\pdfoutput=1
\documentclass[11pt]{article}
\usepackage[final]{acl}
\usepackage{times}
\usepackage{latexsym}
\usepackage{amsmath} 
\usepackage{threeparttable}
\usepackage{multirow} 
\usepackage{booktabs}
\usepackage[T1]{fontenc}
\usepackage[utf8]{inputenc}

\usepackage{microtype}

\usepackage{inconsolata}
\usepackage{url}
\usepackage{graphicx}
\usepackage{color}
\usepackage{xcolor}

\title{Multilingual Gloss-free Sign Language Translation: 

Towards Building a Sign Language Foundation Model}

\author{
  \textbf{Sihan Tan\textsuperscript{1,2}},
  \textbf{Taro Miyazaki\textsuperscript{2}}, 
  \textbf{Kazuhiro Nakadai\textsuperscript{1}}
\\
  \textsuperscript{1}Institute of Science Tokyo,
  \textsuperscript{2}NHK Science \& Technology Research Laboratories
\\
\texttt{\{tansihan, nakadai\} \href{mailto:tansihan@ra.sc.e.titech.ac.jp}{@ra.sc.e.titech.ac.jp}}\\ \href{mailto:miyazaki.t-jw@nhk.or.jp}{\texttt{miyazaki.t-jw@nhk.or.jp}}
  }
\begin{document}
\maketitle
\begin{abstract}
Sign Language Translation (SLT) aims to convert sign language (SL) videos into spoken language text, thereby bridging the communication gap between the sign and the spoken community. While most existing works focus on translating a single sign language into a single spoken language (one-to-one SLT), leveraging multilingual resources could mitigate low-resource issues and enhance accessibility. However, multilingual SLT (MLSLT) remains unexplored due to language conflicts and alignment difficulties across SLs and spoken languages. To address these challenges, we propose a multilingual gloss-free model with dual CTC objectives for token-level SL identification and spoken text generation. Our model supports 10 SLs and handles one-to-one, many-to-one, and many-to-many SLT tasks, achieving competitive performance compared to state-of-the-art methods on three widely adopted benchmarks: multilingual SP-10, PHOENIX14T, and CSL-Daily.\footnote{Codes and model are available:\url{https://github.com/Claire874/Gloss-free-MLSLT}.} 

\end{abstract}

\section{Introduction}

Sign language translation (SLT) is a sophisticated cross-modal task that converts sign language (SL) into spoken language, serving as a crucial bridge between the deaf and hard-of-hearing community and the hearing world. Recent advancements in deep learning have significantly improved SLT performance, particularly through either \textit{gloss-based} or \textit{gloss-free}\footnote{Gloss is another written representation of sign language to help localize sign motions and simplify SLT tasks.} approaches~\cite{camgoz2018neural,chen-two-stream}. While gloss-based methods benefit from intermediate linguistic supervision, they suffer from an information bottleneck, limiting their real-world applicability~\cite{muller-etal-2023-considerations}. In contrast, gloss-free methods directly learn from raw SL videos, making them more practical yet challenging. Despite progress in SLT, existing research predominantly focuses on translating a single SL into a single spoken language (\textit{one-to-one} SLT). However, collecting large-scale annotated SL datasets is difficult. Leveraging multilingual resources could mitigate low-resource issues and enhance accessibility. Existing multilingual SLT (MLSLT) studies~\cite{MLSLT,zhang-etal-2025-improving} are mostly limited to specific datasets (\textit{e.g.,} SP-10) or restricted to \textit{many-to-one} translation. While MLSLT holds great potential, it often suffers from performance degradation due to \textbf{language conflicts}. For instance, we observed a BLEU drop of 1.50 in our universal training setting (In \S~\ref{sec:result} many-to-one). In addition, \textbf{alignment challenges} between SLs and spoken languages hinder the development of MLSLT. To address these limitations, we propose a multilingual gloss-free SLT model with token-level sign language identification (SLI), capable of handling diverse multilingual SLT scenarios. Our contributions are as follows:
\begin{figure}[t]
  \centering
  \includegraphics[width=0.6\columnwidth]{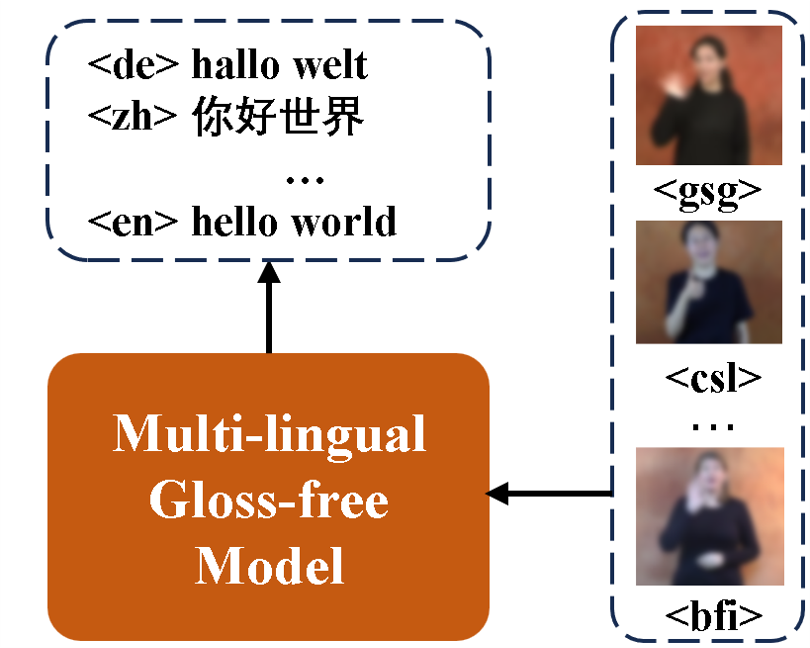}
  \caption{Overview of multilingual gloss-free model. Here, gsg = German Sign Language, csl = Chinese Sign Language, and bfi = British Sign Language.\protect\footnotemark 
 }
  \label{fig:introduction}
  \vspace*{-5mm}
\end{figure}
\footnotetext{We use the ISO639-3 and the ISO693-1 standard to represent sign language and spoken language.}
%To address these limitations, we propose a multilingual gloss-free SLT model that incorporates sign language identification (SLI) and supports one-to-one, many-to-one, and many-to-many SLT. Our contributions are as follows:
\begin{itemize}%\parskip=0pt\itemsep=0pt
    \item We introduce \textit{Sign2(LID+Text)}, a novel SLT approach that adopts dual CTC alignments: one with token-level SL IDs and the other with spoken languages, addressing the language conflicts and alignment challenges in MLSLT.
    
    %\item To the best of our knowledge, we are the first to develop the multilingual gloss-free model supporting 10 SLs and evaluate it on three widely adopted benchmarks.

    %\item We develop a multilingual gloss-free model supporting 10 SLs and including one-to-one, many-to-one, and many-to-many SLT. (emphasize first)
    %\item Our method achieves competitive state-of-the-art performance on PHOENIX14T, CSL-Daily, and multilingual SP-10 dataset. \footnote{Our code and models will be publicly available upon acceptance.}

    %\item To the best of our knowledge, we are the first to explore multilingual gloss-free SLT across three tasks—one-to-one, many-to-one, and many-to-many SLT and evaluate it on three widely adopted benchmark datasets: PHOENIX14T, CSL-Daily, and the multilingual SP-10 dataset. Our model achieves competitive state-of-the-art performance on all three tasks.\footnote{Our code and models will be publicly available upon acceptance.}
    \item To the best of our knowledge, this is the first study to comprehensively explore one-to-one, many-to-one, and many-to-many gloss-free MLSLT, using multiple datasets (SP-10, PHOENIX14T, and CSL-Daily) and achieve state-of-the-art performance for each task.
     % \vspace*{-3mm}
\end{itemize}

\section{Related work}
\label{discussion}
%Our study focuses on identifying and translating multiple sign languages into their corresponding spoken texts. 
%Previous SLT studies can be divided into two categories: \textit{cascading} and \textit{end-to-end}. Cascading SLT, for instance, Sign2Gloss2Text relies on gloss as intermediate supervision. Sign2Gloss2Text simplified SLT by breaking it down into two stages~\cite{yin2020better}: sign language recognition that transcribes a continuous sign video
%to a gloss sequence (Sign2Gloss) and gloss-to-text translation that transforms the glosses into a spoken language text (Gloss2Text). End-to-end SLT instead converts sign language video directly into corresponding spoken text with or without auxiliary supervision.~\citet{camgoz2018neural} propose the Sign2Text task for SLT; however, due to the challenging alignment between sign language and spoken text, it underperformed the co-current cascading SLT. Sign2(Gloss+Text)~\cite{camgoz2020sign} is further proposed, which conducts multi-task SLR and SLT to mitigate the alignment issue. The following research aims to fully explore gloss-free SLT.~\citet{hamidullah-etal-2024-sign} proposed Sign2(Sem+Text), newly introducing sentence embeddings as supervision for gloss-free SLT. Additionally, %Moreover, there have been some attempts at large-scale~\cite{zhang2024scalingsignlanguagetranslation,gueuwou2024shubertselfsupervisedsignlanguage} and multilingual~\cite{MLSLT} SLT. 
Previous SLT studies mainly take a \textit{cascading} or \textit{end-to-end} approach. Cascading SLT, such as Sign2Gloss2Text, introduces gloss as intermediate supervision, simplifying SLT into two stages: sign language recognition (Sign2Gloss) and gloss-to-text translation (Gloss2Text)~\cite{yin2020better}.  
In contrast, end-to-end SLT directly converts sign videos into spoken texts. \citet{camgoz2018neural} first proposed Sign2Text; however, it underperformed cascading SLT due to the challenging sign-text alignment. Later, Sign2Text was integrated with multi-task learning into Sign2(Gloss+Text) to alleviate the alignment issue~\cite{camgoz2020sign}. Recent work further advanced gloss-free SLT. \citet{hamidullah-etal-2024-sign} improved performance by introducing sentence embeddings as supervisions. In addition, large language models (LLMs) opened a new path for gloss-free SLT~\cite{wong2024sign2gpt,gong2024llms,fla-llm}, but their applicability to multilingual settings is limited.

%\paragraph{What is the current status of MLSLT?} MLSLT remains underexplored, hindered by the following challenges. First, SLT itself presents unique difficulties in aligning SL with spoken language. SLs rely on fine-grained articulations such as finger spelling, palm orientation, head movements, and eye aperture~\cite{signlanguagemovement}, which create a tough modality gap. To grasp the linguistic information embedded in sign language, a qualified SLT model must possess both video understanding and sequence generation capabilities~\cite{zhang2023sltunet}. Second, MLSLT struggles with training a unified model across diverse SLs. Some SLs share similarities, whereas others exhibit considerable linguistic and structural differences\footnote{More detailed examples can be found in Appendix~\ref{language conflicts}}~\cite{cross-lingual,jiang-etal-2024-signclip}. These similarities and differences can significantly impact the performance of MLSLT models~\cite{zhang-etal-2025-improving}. Lastly, the scarcity of MLSLT datasets poses a significant challenge to its development. While some studies have explored resource limitations in MLSLT~\cite{MLSLT,gueuwou-etal-2023-jwsign}, research in this area is still in the early stages

\paragraph{What is the current status of MLSLT?}
MLSLT remains underexplored due to several challenges.  
First, SLT itself involves the complex alignment between SL and spoken text, as SLs rely on fine-grained articulations such as finger spelling, palm orientation, and non-manual features~\cite{signlanguagemovement}. Second, language conflicts arise when training a unified model across diverse SLs. While certain SLs exhibit similarities, others differ greatly in structure and lexicon.\footnote{Examples are provided in Appendix~\ref{language conflicts}.}~\cite{cross-lingual,zhang-etal-2025-improving}. An intuitive solution is to introduce utterance-level SLI~\cite{SLI}; however, it is an ill-defined task, as models can learn to identify signers for particular SLs~\cite{jiang-etal-2024-signclip}. Inspired by token-level language identification in multilingual automatic speech recognition (ASR)~\cite{lid}, token-level SLI could offer a more flexible solution. Beyond resolving language conflicts, it could aid the model in mapping SLs to a large multilingual text space by providing fine-grained language cues throughout the sequence. 
Lastly, MLSLT suffers from a lack of large-scale datasets, and despite recent efforts~\cite{MLSLT,gueuwou-etal-2023-jwsign, tanzer2024fleursaslincludingamericansign}, data resources remain scarce.
% Additionally, gloss is a linguistic tool not typically used by deaf and hard-of-hearing individuals for daily communication~\cite{muller-etal-2023-considerations}, further limiting their practical utility in SLT. 
\section{Method}
To address the language conflict and alignment difficulties in MLSLT, we propose \textit{Sign2(LID+Text)}, a novel approach that predicts token-level SL IDs (LID$_\text{tok}$) and translates SLs into spoken languages. Unlike previous studies~\cite{SLI,jiang-etal-2024-signclip} which predict an utterance-level LID  (\emph{e.g.,} a single label <ase> and <en> for American sign language), we introduce two auxiliary CTC objectives~\cite{ctc} to explicitly supervise LID$_\text{tok}$ and target spoken text alignment, enabling hierarchical encoding under a joint CTC/Attention framework, as illustrated in Figure~\ref{fig:introduction}. 
This allows the early encoder layers to focus on token-level SLI, while the later layers reorder the latent sign representations for translation using a text-oriented CTC objective (TxtCTC). Table~\ref{tab: task} summarizes the defined tasks and labels.
\begin{table}[h]
  \centering
   \scalebox{0.85}{ 
  \begin{tabular}{l|ll}

    \hline
    \textbf{Tasks}           & \textbf{Labels}  \\
    \hline
    MLSLT (many-to-many) & $<$\text{en}$>$ hello world                                 \\
    MLSLT (many-to-one) &  hello world                                 \\
    SLT (one-to-one) & hello world\\
    Token-level SLI& $<$\texttt{ase}$>$ $<$\texttt{ase}$>$ $<$\texttt{ase}$>$ \\
  %SLI$_\text{utt}$  & $<$\texttt{ase}$>$ \\

    \hline
  \end{tabular}}
  \caption{Training label examples of Sign2 (LID+Text). In token-level SLI, all tokens are replaced with LIDs, while utterance-level SLI contains a single LID label.}
  \label{tab: task}
    \vspace*{-5mm}
  \end{table}

\begin{figure}[tbh]
  \includegraphics[width=\columnwidth]{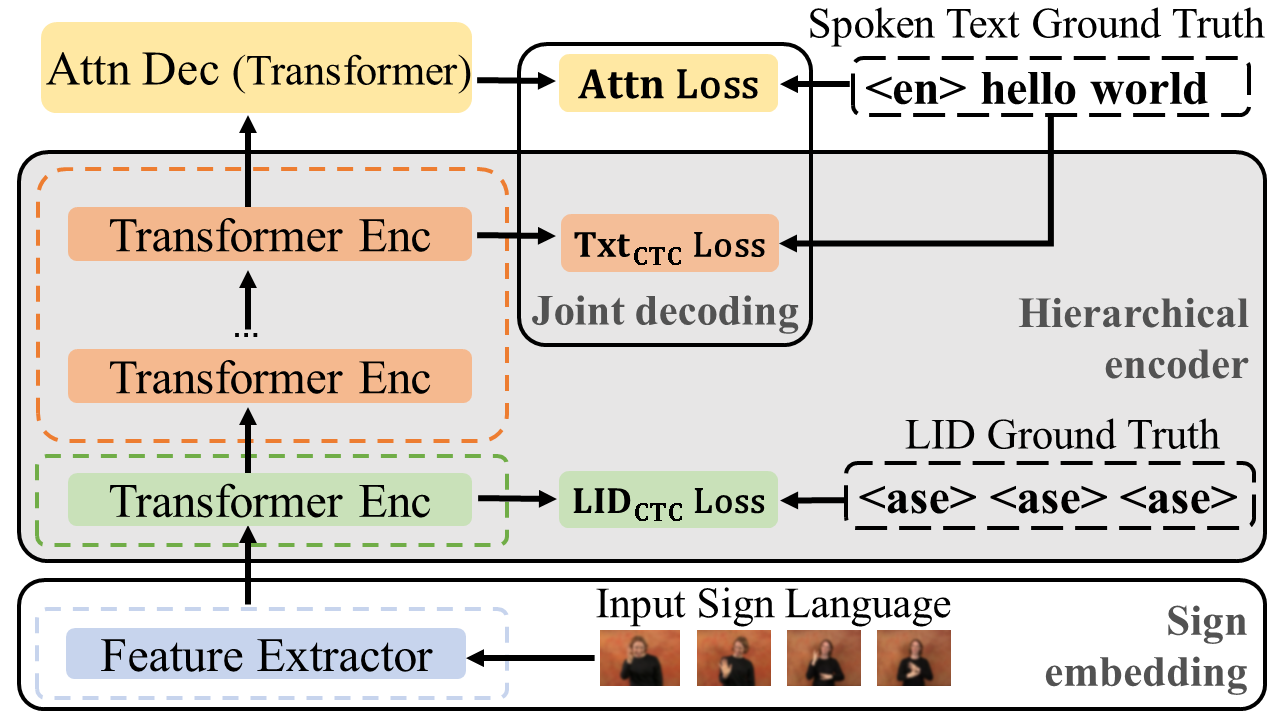}
  \centering
  \caption{Overview of multilingual gloss-free model.}
  \label{fig:introduction}
    \vspace*{-5mm}
\end{figure}
\subsection{Feature extractor}
In previous studies~\cite{zhang2023sltunet,tan-etal-2025-improvement}, a pre-trained feature extractor on glosses has been used as sign embeddings, which is inherently designed for gloss-based SLT. Since our approach is gloss-free, we instead adopt a pre-trained feature extractor based on the SlowFastSign network~\cite{Slowfast} but trained on spoken texts. Pre-trained sign embeddings are further used to extract the sign feature $\mathcal{F}$ from the sign video sequence $\mathcal{V}=\left \{ v_1, v_2,..., v_{|\mathcal{V}|}  \right \} $ consisting of $|\mathcal{V}|$ frames. This process is formulated as:
\begin{equation}
    \mathcal{F} = \text{SignEmbedding}(\mathcal{V}),
    \label{se}
\end{equation}
where $\mathcal{F} = \left \{ f_1, f_2, \dots, f_{|\mathcal{F}|} \right \}$ denotes the extracted feature with $|\mathcal{F}|$ sign representations.

\subsection{Hierarchical encoder}

We employ a CTC-based hierarchical encoder, widely used in ASR~\cite{Hierarchical} and machine translation~\cite{yan-etal-2023-ctc}, to facilitate multi-task learning and improve cross-modal alignments in MLSLT. The hierarchical encoder consists of two modules: an initial token-level SLI (Sign2LID) module and a subsequent Sign2Text module that reorders multilingual sign representations within a joint CTC/Attention framework, optimized with separate CTC objectives.
%CTC-based hierarchical encoding is widely used in ASR~\cite{Hierarchical,lid} and machine translation~\cite{yan-etal-2023-ctc}. In MLSLT, we investigate whether CTC can effectively support multi-task learning and improve alignments or not. To this end, we propose training the early encoder layers with token-level LIDs and the later layers with spoken texts, enabling the model to perform SLI$_\text{tok}$ prediction and reorder multilingual sign representations within a joint CTC/Attention framework.

%CTC-based hierarchical encoding has been widely applied in automatic speech recognition~\cite{Hierarchical,lid} and machine translation~\cite{yan-etal-2023-ctc}. In terms of MLSLT, we aim to investigate a key question: \textit{Can CTC effectively support multi-task learning in MLSLT and improve translation quality?} To explore this, we propose a straightforward approach: training the early and the later encoder layers with different CTC objectives. The early encoder, optimized with LID$_\text{tok}$, performs SLI prediction and adjusts the sign representations length. Meanwhile, the later encoder layer, trained on spoken text, reorders multilingual sign representations, framed within the joint CTC/Attention decoding set-up. 
\paragraph{Sign2LID module} is to predict the LID$_\text{tok}$, $\mathcal{I}_\text{tok} = \left \{ i_1, i_2,..., i_{|\mathcal{T}|}  \right \} $ with the same length as the spoken text $\mathcal{T} = \left \{ t_1, t_2,..., t_{|\mathcal{T}|}  \right \} $. 
 We incorporate the LID$_\text{tok}$-oriented CTC loss as part of the multi-task Sign2(LID+Text) objective function.
\begin{equation}
  \label{eq:LID CTC Loss}
  \mathcal{L}_\text{LID} = -\text{log}P_\text{CTC}(\mathcal{I}_\text{tok}|\mathcal{F}).
\end{equation}
As in hierarchical conditioning, deeper layers handle increasingly complex predictions~\cite{higuchi2022hierarchical}; the initial encoder layer suffices for the Sign2LID task. We assign this auxiliary task to the initial encoder layer, where the $\mathcal{|T|}$-length LID$_\text{tok}$ sequence explicitly aligns the sign representations with each spoken word as
\begin{equation}
  \label{eq:hint}
  \textbf{h}_{\text{int}} = \textbf{Enc}_\text{int}(\mathcal{F}).
\end{equation}
The intermediate sign representations $\textbf{h}_\text{int}$ from the initial encoder layer, $\textbf{Enc}_\text{int}$, are then forwarded to the subsequent encoder layers for Sign2Text.
%This alignment enables the subsequent encoder layers to learn more effectively for MLSLT. (need say more?)
\paragraph{Sign2Text module} reorders the sign representations into the spoken text sequence. While CTC is typically constrained to monotonic alignments, neural network encoders allow for latent reordering~\cite{zhang2022revisiting}, enabling CTC to handle the non-monotonic alignment between the SL and spoken text. We apply TxtCTC to align the final encoder representation $\textbf{h}_{\text{fin}}$ with the target spoken text sequence $\mathcal{T}$:
\begin{equation}
  \label{eq:LID CTC Loss}
  \mathcal{L}_\text{Txt} = -\log P_\text{CTC}(\mathcal{T}|\textbf{h}_{\text{fin}}).
\end{equation}

Following previous work~\cite{yan-etal-2023-ctc}, we frame our decoding process within a joint CTC/Attention setup, where the attention decoder plays a leading role in generating the output sequence, and the TxtCTC score provides auxiliary guidance during beam search.  
The overall training objective function jointly optimizes the hierarchical encoder and the attention decoder:
\begin{equation}
  \label{eq:total Loss}
  \mathcal{L}_\text{total} = \lambda_1\mathcal{L}_\text{LID} + \lambda_2\mathcal{L}_\text{Txt} + \lambda_3\mathcal{L}_\text{Attn},
\end{equation}
where $\mathcal{L}_\text{Attn}$ denotes the maximum likelihood estimation (MLE) loss for MLSLT, and $\lambda$s control contributions of the Sign2LID, Sign2Text, and attention decoder objectives.

\section{Experimental settings}
To validate our proposed method, we conduct experiments on three tasks: one-to-one, many-to-one, and many-to-many SLT.  % Each task aims to verify the proposed components, respectively. \textbf{One-to-one:} By definition, it is a task that converts one sign language into its corresponding spoken text. Most SLT studies focus on one-to-one SLT. \textbf{Many-to-one:} This task translates multiple sign languages into one specific spoken text. \textbf{Many-to-many:} This is the most challenging situation, where the model needs to translate multiple sign languages into multiple spoken languages. 
\paragraph{Datasets.}
We utilize three widely adopted datasets for our experiments: the multilingual SP-10~\cite{MLSLT}, RWTH-PHOENIX-2014T (PHOENIX14T)~\cite{camgoz2018neural}, and CSL-Daily~\cite{zhou2021improving}. SP-10 supports a broader range of tasks, featuring video recordings of 10 SLs from
SpreadTheSign~\cite{hilzensauer2015multilingual}. In contrast, PHOENIX14T and CSL-Daily are designed for one-to-one SLT. Appendix \ref{sec:SP-10information} provides statistics for the three datasets.

\paragraph{Implementation Details.}  
We adopt a Transformer-based architecture within a joint CTC/Attention decoding framework~\cite{tan-etal-2025-improvement}. To evaluate the effectiveness of our method, we compare it with a vanilla Transformer baseline~\cite{vaswani2017attention}. Full implementation details and hyperparameters are provided in Appendix~\ref{sec:ImplementDetail}.

%We implement our model using the Transformer architecture~\cite{vaswani2017attention} within a joint CTC/Attention decoding framework~\cite{yan-etal-2023-ctc}. We use a vanilla Transformer as our baseline for performance comparison. Full implementation details, including hyperparameters and training settings, are provided in Appendix~\ref{sec:appendix}.
%The model is trained on an NVIDIA A100 (80GB) GPU with a batch size of 64. The training objectives \(\lambda_1\), \(\lambda_2\), and \(\lambda_3\) are set to 1, 5, and 3, respectively. If \(\lambda_1\) is not involved in the training, it is set to 0.%The model is configured with a hidden size of 256 and a feed-forward dimension of 2048. Both the encoder and decoder consist of six layers. Training is conducted using the Xavier initializer and the Adam optimizer~\cite{adam} with a learning rate of \(1 \times 10^{-3}\). 

\paragraph{Evaluation metrics.} Following the previous studies, we evaluate performance using BLEU~\cite{papineni2002bleu} and ROUGE~\cite{lin2004rouge}. BLEU is calculated through SacreBLEU~\cite{post-2018-call}.
\section{Results and Discussion}
\label{sec:result}
\begin{figure}[h]
  \centering
  \includegraphics[width=\columnwidth]{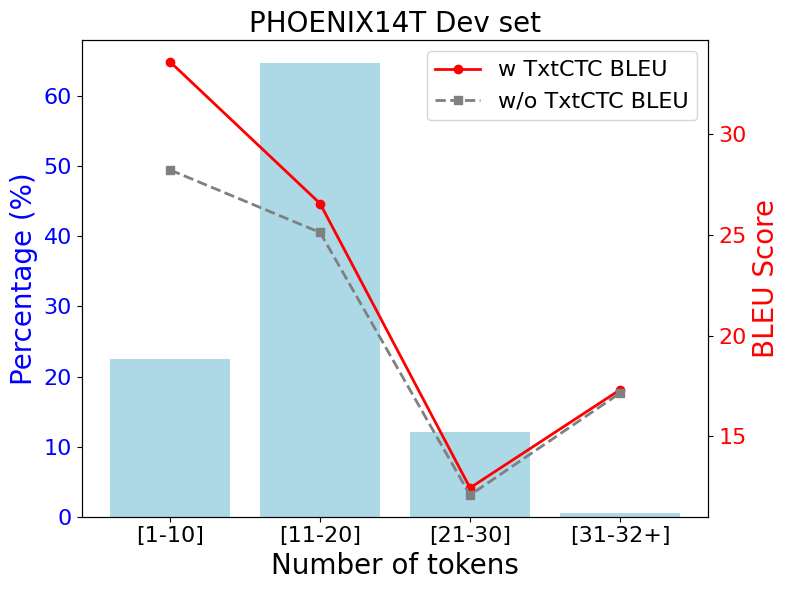}
  \caption{Average BLEU score on different token length intervals on PHOENIX14T.}
  \label{fig:phoenix}
    \vspace*{-5mm}
\end{figure}

\begin{figure}[h]
  \centering
  \includegraphics[width=\columnwidth]{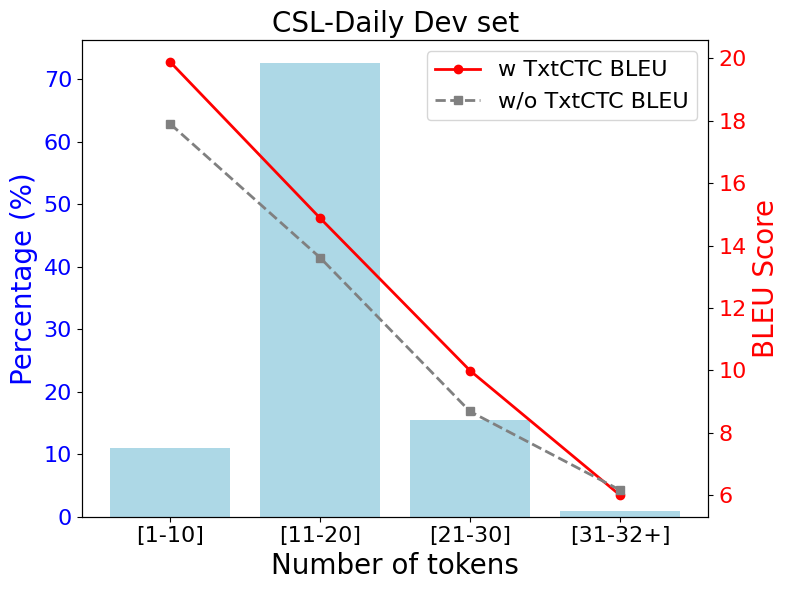}
  \caption{Average BLEU score on different token length intervals on CSL-Daily.}
  \label{fig:csl}
    \vspace*{-5mm}
\end{figure}

\paragraph{One-to-one} evaluates the alignment capability of TxtCTC in standard single-pair SLT. Since Sign2LID is not utilized in this task, it is a purely gloss-free SLT setting. Tables~\ref{tab: one-to-one of SP-10} and~\ref{table: one2one gloss-free} present results on SP-10, PHOENIX14T, and CSL-Daily, respectively. Our TxtCTC, integrated within the joint CTC/Attention framework, achieves 1.71 and 2.24 BLEU improvements on PHOENIX14T and CSL-Daily. To further investigate the effect of TxtCTC, we present the token length distribution of PHOENIX14T and CSL-Daily along with the average BLEU4 score on each interval (see Figures~\ref{fig:phoenix} and~\ref{fig:csl}). We observed that proposed TxtCTC within the joint CTC/Attention framework tends to be more effective for short and medium-length sentences. The impact of TxtCTC diminishes as sentence length increases.
CTC tends to be more effective for segments with fewer ambiguities and clearer frame-to-token correspondences, which are more common in shorter sequences. For longer sequences, the model relies more on the attention mechanism to capture global context, which may naturally reduce the marginal contribution of the TxtCTC objective.
In addition, Gloss-free SLT introduces an additional challenge that the input SL frame sequence is typically much longer than the spoken sentence. This inherent length difference increases the alignment difficulty, particularly for long sentences. 

As no prior work reports one-to-one SLT results on SP-10 beyond English, we provide the first benchmark to facilitate future research.
  \begin{table}[h]
  \centering
  \scalebox{0.8}{
  \begin{tabular}{lcccc}
    \hline
    \multirow{2}{*}{\textbf{Language Pairs}} & \multicolumn{2}{c}{\textbf{Dev}} & \multicolumn{2}{c}{\textbf{Test}}   \\
    \cmidrule(lr){2-3}\cmidrule(lr){4-5}
    & BLEU& ROUGE& BLEU& ROUGE\\
    \hline
    csl $\to$ zh &8.79&35.59&7.32&32.40 \\
    ukl $\to$ uk & 7.47 &32.56&  6.84&30.12 \\   
    rsl $\to$ ru & 6.20&31.79&4.23&28.98 \\
    icl $\to$ is&4.79&27.57&4.25&30.45\\
    gsg $\to$ de& 6.07& 33.73&5.77&32.20\\
    ise $\to$ it&5.88&30.13&4.76&27.91\\
    bqn$\to$bg&4.77 &  28.93 &2.59&23.91\\
    swl $\to$ sv& 7.47 &31.31&7.23 &30.45\\   
    lls $\to$ lt& 2.33&26.70&2.42&24.36\\
    bfi $\to$ en&7.80&33.77&6.23&32.33\\
    
    \hline
  \end{tabular}}
  \caption{One-to-one SLT results on the SP-10 dataset.}
  \label{tab: one-to-one of SP-10}
    \vspace*{-5mm}
  \end{table}
%The key exploration is to verify the effectiveness of spoken-text CTC with a joint CTC/Attention decoding setting.         
\begin{table*}[h]
\centering
\begin{threeparttable}
\scalebox{0.85}{
\begin{tabular}{lcccccccc}%四个c代表有四列且内容居中
\toprule%第一道横线
\multirow{3}{*}{\textbf{Methods}}&\multicolumn{4}{c}{\textbf{PHOENIX14T}} &\multicolumn{4}{c}{\textbf{CSL-Daily}} \\
%\cmidrule(lr){2-3}\cmidrule(lr){4-5} \cmidrule(lr){6-7}
%\midrule%第二道横线 %跨两列、内容居中、跨列内容为Resultsummary
  &\multicolumn{2}{c}{\underline{}\textbf{Dev}} &\multicolumn{2}{c}{\textbf{Test}}&\multicolumn{2}{c}{\textbf{Dev}} &\multicolumn{2}{c}{\textbf{Test}}\\
   \cmidrule(lr){2-3}\cmidrule(lr){4-5} \cmidrule(lr){6-7} \cmidrule(lr){8-9}
  & BLEU& ROUGE& BLEU& ROUGE& BLEU& ROUGE& BLEU& ROUGE\\
\hline
\textbf{\textit{Gloss-free}}&&&&&&&&\\

NSLT+Luong~\cite{camgoz2018neural}& 10.00 & 32.60&9.00& 30.70& 7.96&34.28&7.56&34.54\\
CSGCR~\cite{CSGCR}&15.08&38.96&15.18&38.85&--&--&--&--\\
GFSLT-VLP~\cite{GFSLT-VLP}&22.12&43.72&21.44&42.49&11.07&36.70& 11.00&36.44\\
Sign2GPT~\cite{wong2024sign2gpt}&--&--& 22.52&48.90&--&--& 15.40& 42.36\\
Fla-LLM~\cite{fla-llm}&--&--& 23.09&45.27&--&--&14.20&37.25\\
SignLLM~\cite{gong2024llms}&\textbf{25.25}&47.23&23.40&44.49&12.23&39.18&\textbf{15.75}&39.91\\
\hline

\textit{\textbf{Baseline}}&22.59&49.88&22.52&49.85&12.23&36.39&11.76&36.25\\ 
\textit{\textbf{Ours \textbf{w} TxtCTC}}&24.18&\textbf{51.74}&\textbf{24.23}&\textbf{50.60}&\textbf{13.66}&\textbf{39.33}&14.18&\textbf{40.00}\\

\bottomrule%第四道横线
\end{tabular}} 
\end{threeparttable}

\caption{Experimental results on PHOENIX14T and CSL-Daily dataset for gloss-free SLT (one-to-one SLT).}
\label{table: one2one gloss-free}
%\vspace*{-5mm}
\end{table*}

\paragraph{Many-to-one}
\begin{table*}
  \centering
  \scalebox{0.7}{
  \begin{tabular}{c|lcccccccccccc}
   \toprule
    \textbf{Part/Metrics} & \textbf{Methods}& \textbf{csl}&\textbf{ukl}&\textbf{rsl}&\textbf{bqn}&\textbf{icl}&\textbf{gsg}&\textbf{ise}&\textbf{swl}&\textbf{lls}&\textbf{bfi}&\textbf{Mean}\\%&\textbf{All} \\
    \hline
   & Individual (10)&\textbf{8.02}&\underline{6.32}&\underline{5.56}&\textbf{4.88}&5.03&\underline{5.54}&4.78&\underline{7.54}&\underline{5.15}&\underline{6.42}&\underline{5.92}\\
    Dev&Universal (1)&5.93&5.32&4.91&3.15&4.69&4.65&4.52&6.63&4.47&4.59&4.89\\
   /& Google Multi~\cite{googles}&2.46& 3.14& 2.93& 2.21& 3.44& 2.71 &3.18 &2.89 &1.81& 3.49 &2.83\\
   BLEU & MLST~\cite{MLSLT}& 5.16&5.42&4.95&3.28&\textbf{6.76}&5.18&\textbf{7.05}&6.33&\textbf{6.08}&7.03&5.72\\
    % Dev& Individual (10)&8.07&&&&&&&&&&\\
   &  \textbf{\textit{Ours}}&\underline{7.06}&\textbf{6.77}&\textbf{7.38}&\underline{3.56}&\underline{6.59}&\textbf{5.59}&\underline{4.83}&\textbf{7.76}&4.54&\textbf{7.79}&\textbf{6.19}\\
    %BLEU&  w Sign2(Lid+Text)&\\
    \hline
    & Individual (10)&\underline{36.60}&33.61&30.05&\underline{27.70}&31.51&32.47&29.88&35.80&\underline{31.44}&32.57&28.53\\
    Dev&Universal (1)&32.74&31.64&31.37&26.19&30.18&29.26&30.12&33.19&31.16&30.22&30.61\\
    /& Google multi~\cite{googles}&28.50& 28.93& 30.01& 24.66& 29.91& 29.75& 28.33& 31.01 &27.7 &32.42& 29.12\\
    ROUGE&MLSLT~\cite{MLSLT}&34.59&\underline{34.04}&\underline{31.62}&\textbf{27.98}&\underline{35.29}&\underline{33.50}&\textbf{37.96}&\underline{36.02}&\textbf{34.48}&\underline{37.25}&\underline{34.27}\\
    %Dev& Individual (10)&34.42\\
    & \textbf{\textit{Ours}}&\textbf{36.77}&\textbf{34.86}&\textbf{36.02}&26.74&\textbf{35.64}&\textbf{34.95}&\underline{31.67}&\textbf{37.97}&31.43&\textbf{37.56}&\textbf{34.36}\\
    %ROUGE& w Sign2(Lid+Text)\\
    \hline
    & Individual (10) &\textbf{6.24}&4.00&\underline{3.69}&\textbf{3.63}&3.30&3.77&3.40&\underline{6.21}&4.49&\underline{6.23}&4.60\\
    Test&Universal (1)&2.72&2.36&2.19&\underline{3.02}&\underline{4.14}&3.31&1.19&3.94&3.20&4.70&3.10\\
    /&Google Multi~\cite{googles}&2.28& 2.38& 2.06 &1.10& 1.38& 1.82& 2.09& 2.13 &2.68& 3.27& 2.12\\
   BLEU& MLSLT~\cite{MLSLT}&5.19&\underline{4.18} &3.66 &2.85& 3.93 &\textbf{4.97}& \textbf{6.70} &3.70& \textbf{5.72} &5.73 &\underline{4.66}\\
    %Test& Individual (10)&4.84&\\
    &\textbf{\textit{Ours}}&\underline{5.92}&\textbf{4.52}&\textbf{5.80}&2.93&\textbf{5.10}&\underline{4.65}&\underline{5.00}&\textbf{6.40}&\underline{5.13}&\textbf{6.36}&\textbf{5.18}\\
     %BLEU&w Sign2(Lid+Text)\\
     \hline
     & Individual (10)&\underline{34.51}&30.93&30.90&\textbf{27.19}&28.00&30.19&27.66&\underline{33.92}&31.36&\underline{35.73}&31.04\\
    Test&Universal (1)&29.57&29.03&28.96&\underline{26.66}&31.75&30.12&26.73&31.11&30.33&32.41&29.67\\
     /&Google multi~\cite{googles}& 29.37& 28.63& 29.57& 23.95& 28.53& 29.36& 29.30& 29.83 &30.03 &30.76& 28.93\\
     ROUGE&MLSLT~\cite{MLSLT}&33.33 &\textbf{34.07} &\underline{31.54} &25.75 &\underline{33.25}& \underline{32.13} &\textbf{35.37}&33.09 &\textbf{33.11}&35.34& \underline{32.70}\\
    % Test& Individual (10)&28.66\\
     &\textbf{\textit{Ours}}&\textbf{35.18}&\underline{33.17}&\textbf{34.17}&24.55&\textbf{34.57}&\textbf{32.83}&\underline{34.27}&\textbf{34.98}&\underline{31.48}&\textbf{36.16}&\textbf{33.14}\\
     %ROUGE& w Sign2(Lid+Text)\\
    \bottomrule
  \end{tabular}}
  \caption{Many-to-one SLT results on the SP-10 dataset, we select English as the target spoken language. The best performance is bolded, and the second-best is underlined.}
  \label{tab: many-to-one of SP-10}
    \vspace*{-5mm}
  \end{table*}
  evaluates many-to-one SLT on the SP-10 dataset. Following previous studies~\cite{MLSLT,zhang-etal-2025-improving}, we selected English as the target spoken language. A major challenge in the many-to-one setting is the language conflict, as confirmed by a preliminary experiment: the baseline many-to-one model suffers an average BLEU drop of 1.50 compared to individually trained one-to-one models (see Table~\ref{tab: many-to-one of SP-10} individual(10) and universal (1)). Instead, our \textit{Sign2(LID+Text)} approach mitigates the language conflict and surpasses individual translation by 0.58 BLEU. Table~\ref{tab: many-to-one of SP-10} reports the results of each SL. Overall, our model outperforms previous MLSLT systems. However, the limited target vocabulary ($\sim$1.1k words) constrains further improvements. Data augmentation could be a promising way to address this limitation.

%\paragraph{Many-to-many} is the most challenging setting. We incrementally add language pairs based on their performance in Table~\ref{tab: one-to-one of SP-10}, from highest to lowest on the dev set. Table~\ref{tab:Many-to-many of SP-10} presents the comparison with one-to-one SLT. Our many-to-many model maintains comparable performance as the number of language pairs increases to five. This stability may benefit from cross-lingual information sharing, which reduces the need for large-scale data, particularly in low-resource SLT scenarios. To further verify the effectiveness of LID$\text{tok}$ in MLSLT, we conduct an ablation study (Appendix \ref{sec: ablation study of token-level LID}), which shows that LID$\text{tok}$ is especially effective in more challenging translation settings. Many-to-many models offer a promising, scalable solution to bridge the data gap toward building an SL foundation model. %However, as the number of languages increases, the risk of cross-lingual interference may also grow, posing a potential trade-off.

\paragraph{Many-to-many} is the most challenging setting. We incrementally add language pairs based on their performance in Table~\ref{tab: one-to-one of SP-10}, from highest to lowest on the dev set (see Appendix~\ref{Details of language pairs}). Table~\ref{tab:Many-to-many of SP-10} presents the comparison with one-to-one SLT. Our many-to-many model maintains comparable performance as the number of language pairs increases to five. This stability benefits from cross-lingual information sharing, which reduces the reliance on large-scale data, particularly in low-resource SLT scenarios. To further validate LID$\text{tok}$, we conduct an ablation study (see Appendix \ref{sec: ablation study of token-level LID}) and find that LID$\text{tok}$ is especially effective under more challenging translation conditions. These results suggest that our many-to-many model provides a promising and scalable solution to mitigating the data scarcity problem and paves the way toward a unified SL foundation model in challenging SLT settings.%These results suggest that many-to-many models offer a promising and scalable approach to bridging the data gap and building an SL foundation model.

\begin{table}[h]
  %\centering
  \scalebox{0.8}{
  \begin{tabular}{lcccc}
    \hline
   \multirow{2}{*}{\textbf{Language Pairs}} & \multicolumn{2}{c}{\textbf{One-to-one}} & \multicolumn{2}{c}{\textbf{Many-to-many}}   \\
   % \textbf{Language Pairs} & BLEU& ROUGE \\
    \cmidrule(lr){2-3}\cmidrule(lr){4-5}
    & BLEU& ROUGE& BLEU& ROUGE\\

    \hline
    (2$\to$2) &6.28 &32.37&6.22&34.53\\
    (3$\to$3) & 6.44&31.73&6.15&34.02 \\   
    (4$\to$4) &6.41&34.59&6.31 & 32.76\\
    (5$\to$5)&6.33&33.88&5.36&31.79\\
    (6$\to$6)&6.08&33.40 &4.91&31.17\\
    (7$\to$7)&5.97&34.37&4.63&29.51\\
    (8$\to$8)&5.84&32.70&4.75&30.24  \\
   (9$\to$9)&5.27&32.14& 4.74 &28.65\\   
    (10$\to$10)&5.06&32.31&4.58&30.83\\
    %MLSLT~\cite{MLSLT}&1.88 &30.26 \\

    \hline
  \end{tabular}}
  \caption{One-to-one vs. many-to-many SLT.}
  \label{tab:Many-to-many of SP-10}
  \vspace*{-5mm}
  \end{table}

\section{Conclusion}
To address language conflicts and alignment challenges in multilingual sign language translation (MLSLT), we proposed \textit{Sign2(LID+Text)}, a multilingual gloss-free SLT model combining token-level sign language identification (Sign2LID) and sign-to-text CTC alignment (Sign2Text). Our approach achieved comparable performance with the state-of-the-art across one-to-one, many-to-one, and many-to-many SLT tasks on three widely adopted benchmarks, covering a total of 10 different sign languages (SLs). We showed that Sign2LID effectively mitigates language conflicts and Sign2Text improves sign-to-text alignment, especially for shorter and medium-length sequences. Our work encourages and lays the foundation for future exploration of large-scale multilingual gloss-free SLT and shows potential for enhancing cross-lingual SL processing, contributing to the development of a universal SL foundation model.%Our work encourages and lays the foundation for future exploration of large-scale multilingual gloss-free SLT. Our method shows potential for enhancing cross-lingual SL processing and advancing the development of an SL foundation model.
%We develop a multilingual gloss-free sign language translation (SLT) model by proposing \textit{Sign2(LID+Text)}, which leverages CTC alignments with token-level language IDs and spoken language to address language conflicts and alignment challenges in MLSLT. Our model achieves competitive performance on three benchmark datasets (PHOENIX14T, CSL-Daily, SP-10) across one-to-one, many-to-one, and many-to-many SLT tasks, supporting translation among 10 sign languages. 
%Furthermore, our many-to-many task analysis suggests that multilingual modeling can compensate for data scarcity in low-resource SLs and holds promise for practical applications. For future work, we aim to scale the model to a broader range of sign languages, conduct more challenging settings, and further explore cross-lingual transfer and pretraining toward a foundation sign language model.

%Our work lays the foundation for future exploration of large-scale multilingual gloss-free SLT and shows potential for enhancing cross-lingual SL processing, contributing to the development of a universal SL foundation model.

\section*{Limitations}
The limitations of this work can be summarized as follows.
First, data scarcity remains a major challenge. The SP-10 dataset is currently the only publicly available multilingual SLT corpus, and as shown in Appendix~\ref{sec:SP-10information}, each language in SP-10 contains only 830 training samples (8.3k overall), which is extremely small for training deep learning models. Moreover, as discussed in the many-to-one setting, the target language vocabulary is limited to approximately 1.1k words, further constraining the model’s capacity to generate diverse outputs. Second, our many-to-many SLT evaluation is set to, at most, 10 language pairs. Extending the evaluation to the full 10$\times$10 combinations poses a greater challenge and requires more computational resources. Future work will focus on scaling to more sign languages and more challenging settings.

%\section*{Ethical Considerations}
%This study is based on publicly available datasets (PHOENIX14T, CSL-Daily, SP-10) and adheres to original licensing and usage agreements.  
%We acknowledge that sign languages are the natural languages of Deaf and hard-of-hearing communities, and the development of SLT systems should be conducted with respect for their linguistic and cultural integrity.  
%While our work focuses on model development, we recognize the critical importance of involving deaf and hard-of-hearing researchers, sign language users, and linguists in the research and evaluation process to ensure that SLT technologies align with their communication needs and social values.
\section*{Acknowledgment}
We thank the (meta-)reviewers for their valuable feedback.

\bibliography{custom}
\newpage
\appendix

%\section{Appendix}
%\label{sec:appendix}

%\paragraph{Examples of sign language conflicts}
\section{Examples of sign language conflict}
\label{language conflicts}
\begin{figure}[thb]
  \centering
  \includegraphics[width=\columnwidth]{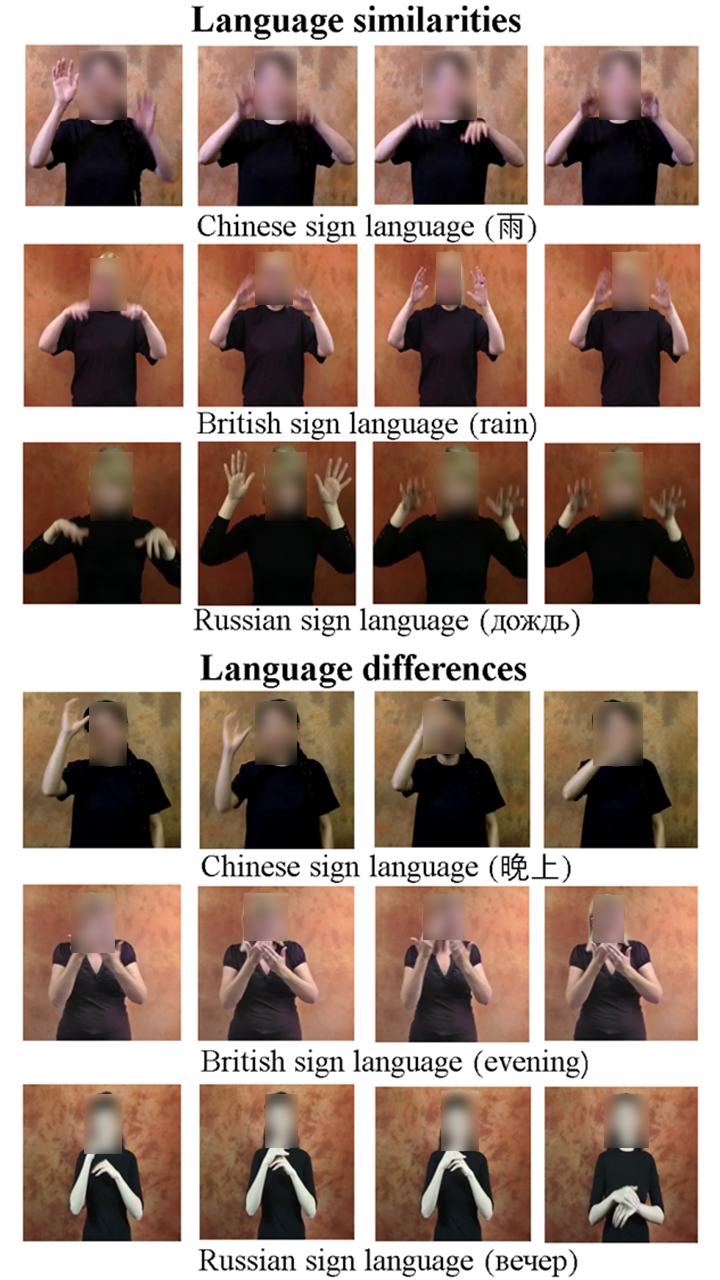} % Fixed filename issue
  \caption{Sign language similarities and differences across languages. Sign videos are from SpreadTheSign. Note for privacy: we anonymize signers.}%\protect\footnotemark}
  %\footnotetext{SpreadTheSign:\protect\url{https://www.spreadthesign.com/en.us/search/?}}
  \label{fig:language_similarities} % Fixed typo in label
\end{figure}

As shown in Figure~\ref{fig:language_similarities}, many sign languages share similarities in expressing certain concepts. For example, when signing \textit{rain}, signers often mimic the shape of raindrops falling, which is relatively universal. However, for \textit{evening}, although the core concept involves representing the sun setting, variations in expression still exist across different sign languages.
%\footnotetext{SpreadTheSign:\protect\url{https://www.spreadthesign.com/en.us/search/?}}

%\paragraph{Implement details}
\section{Implement details}
\label{sec:ImplementDetail}
The Transformer model with a CTC/Attention setup uses a hidden size of 256 and a feed-forward dimension of 2048. Both the encoder and decoder have six layers. Training is conducted using the Xavier initializer and the Adam optimizer~\cite{adam} with a learning rate of \(1 \times 10^{-3}\). We train the model on an NVIDIA A100 (80GB) GPU with a batch size of 64. The total trainable \#params is 39.48M. The training objective weights are set as \(\lambda_1 = 1\), \(\lambda_2 = 5\), and \(\lambda_3 = 3\). When token-level LID is not used, \(\lambda_1\)  is set to 0. 

%\paragraph{Language information in SP-10} 
\section{ Summary of different SLT datasets}
\label{sec:SP-10information}
We summarize the statistics of the different datasets in Table~\ref{tab: details_of_3_dataset}, SP-10 consists of 10 different sign languages, with each having 830 training samples. In addition, Table~\ref{tab: details of SP-10} shows the involved sign languages with their abbreviations in SP-10.
\begin{table}[tbh]
\scalebox{0.65}{
\begin{tabular}{lcccccc}
\hline
\multirow{2}{*}{\textbf{Datasets}} & \multirow{2}{*}{\textbf{Lang}} & \multicolumn{5}{c}{\textbf{Statistics}} \\
\cmidrule(lr){3-7}
& & \#Signer & Vocab & \#Train & \#Dev & \#Test \\
\hline
SP-10      & multilingual & 79 & 16.7k & 8,300 & 1,420 & 2,021 \\
PHOENIX14T & gsg & 9 & 2.9k & 7,096 & 519 & 642 \\
CSL-Daily  & csl & 10 & 2.3k & 18,401 & 1,077 & 1,176 \\
\hline
\end{tabular}}
\caption{Statistics of SP-10, PHOENIX14T, and CSL-Daily datasets.}
\label{tab: details_of_3_dataset}
  \vspace*{-5mm}
\end{table}

\begin{table}[h]
  \centering
  \scalebox{0.95}{
  \begin{tabular}{ll}
    \hline
   \textbf{Languages}           & \textbf{Abbr.}   \\
    \hline
   Chinese sign language & csl                              \\
    Ukrainian sign language& ukl\\   
    Russian sign language & rsl \\
    Icelandic sign language & icl\\
    German sign language & gsg\\
    Italian sign language & ise\\
    
Bulgarian sign language& bqn\\
    Swedish sign language & swl\\   
Lithuanian sign language & lls\\
 
British sign language& bfi\\
    \hline
  \end{tabular}}
  \caption{ Sign language abbreviations of the SP-10 dataset.}
  \label{tab: details of SP-10}
  \vspace*{-5mm}
  \end{table}

%\paragraph{Language conflicts in SP-100}
\section{The language conflict in SP-10}
\label{sec:conflict_SP100}
We performed a preliminary experiment to investigate language conflicts in multilingual SLT. The baseline model is adopted for individual (10) and universal (1) many-to-one SLT. The individual and universal translation performances are presented in Table~\ref{tab: Sign language conflicts}. In general, the universal has an average 1.50 BLEU performance drop. 
  \begin{table}[h]
  \scalebox{0.7}{
  \begin{tabular}{lccc}
    \hline
    \multirow{2}{*}{\textbf{Language Pairs}} & \multicolumn{1}{c}{\textbf{Individual (10)}} & \multicolumn{1}{c}{\textbf{Universal (1)}}&\multirow{2}{*}{\textbf{Variation}}   \\
    %\cmidrule(lr){2}\cmidrule(lr){4-5}
    & BLEU& BLEU\\
    \hline
    csl $\to$en&6.24&2.72&$-$3.52  \\
    ukl $\to$ en& 4.00&2.36 &$-$1.64\\   
    rsl $\to$ en&3.69& 2.19&$-$1.50\\
    icl $\to$ en&3.30&4.14&$+$0.84\\
    gsg $\to$ en& 3.77&3.31&$-$0.46\\
    ise $\to$ en&3.40&1.19&$-$2.21\\
    bqn$\to$en&3.63&3.02&$-$0.61\\
    swl $\to$ en& 6.21&3.94&$-$2.27\\   
    lls $\to$ en&4.49&3.20& $-$1.29\\
    bfi $\to$ en&6.23 &4.70&$-$1.53\\
   \textbf{ Mean}&\textbf{4.60}&\textbf{3.10}&\textbf{$-$1.50}\\
    \hline
  \end{tabular}}
  \caption{Language conflicts in SP-10, we present the individual and universal translation results on the baseline.}
  \label{tab: Sign language conflicts}
    %\vspace*{-5mm}
  \end{table} 

\section{Ablation study of token-level LID}
\label{sec: ablation study of token-level LID}
Performance deteriorates as the number of languages involved in the many-to-many translation increases, and the translation task becomes more complex. As shown in Table~\ref{tab:ablation_study_of_LID}, our method using token-level LID can suppress this deterioration and is effective in more complex translation settings.
\begin{table}[tbh]
  \centering
  \scalebox{1}{
  \begin{tabular}{lcc}
    \hline
    \multirow{2}{*}{\textbf{Language Pairs}} & \textbf{w/o LID$_\text{tok}$} & \textbf{w LID$_\text{tok}$} \\
    & BLEU & BLEU \\
    \hline
    (2$\to$2)  & \textbf{7.48}&6.22  \\
    (3$\to$3)  &\textbf{6.50}&6.15 \\   
    (4$\to$4)  &4.98&\textbf{6.31} \\
    (5$\to$5)  & 4.74&\textbf{5.36 } \\
    (6$\to$6)  &4.57&\textbf{ 4.91} \\
    (7$\to$7)  &3.70&\textbf{4.63}  \\
    (8$\to$8)  &4.27&\textbf{4.75}  \\
    (9$\to$9)  &3.56&\textbf{4.74} \\   
    (10$\to$10)&3.87&\textbf{4.58}  \\
    \hline
  \end{tabular}}
  \caption{Ablation study of token-level language ID (LID$_\text{tok}$) in many-to-many SLT.}
  \label{tab:ablation_study_of_LID}
\end{table}

\section{Details of language pairs in many-to-many SLT}
\label{Details of language pairs}
Table~\ref{tab:Details of language pairs} shows each language pair used in our many-to-many translation experiment.
\begin{table}[h]
  \centering
  \scalebox{0.9}{
  \begin{tabular}{ll}
    \hline
    \textbf{Language Pairs}  \\
    \hline
    (2$\to$2)  & (csl$\to$zh) (bfi$\to$en) \\
    (3$\to$3)  & (csl$\to$zh) (bfi$\to$en) (swl$\to$sv)\\   
    (4$\to$4)  & (csl$\to$zh) (bfi$\to$en) (swl$\to$sv) (ukl$\to$uk)\\
    (5$\to$5)  & (csl$\to$zh) (bfi$\to$en) (swl$\to$sv) (ukl$\to$uk) (gsg$\to$de)\\
    (6$\to$6)  & (csl$\to$zh) (bfi$\to$en) (swl$\to$sv) (ukl$\to$uk) (gsg$\to$de) (ise$\to$it)\\
    (7$\to$7)  & (csl$\to$zh) (bfi$\to$en) (swl$\to$sv) (ukl$\to$uk) (gsg$\to$de) (ise$\to$it) (rsl$\to$ru) \\
    (8$\to$8)  & (csl$\to$zh) (bfi$\to$en) (swl$\to$sv) (ukl$\to$uk) (gsg$\to$de) (ise$\to$it) (rsl$\to$ru) (icl$\to$is)\\
    (9$\to$9)  & (csl$\to$zh) (bfi$\to$en) (swl$\to$sv) (ukl$\to$uk) (gsg$\to$de) (ise$\to$it) (rsl$\to$ru) (icl$\to$is) (bqn$\to$bg)\\   
    (10$\to$10) & (csl$\to$zh) (bfi$\to$en) (swl$\to$sv) (ukl$\to$uk) (gsg$\to$de) (ise$\to$it) (rsl$\to$ru) (icl$\to$is) (bqn$\to$bg) (lls$\to$lt) \\
    \hline
  \end{tabular}}
  \caption{Language pairs used in many-to-many SLT}
  \label{tab:Details of language pairs}
  \end{table}

\end{document}